\documentclass[conference]{ieeeconf}
\IEEEoverridecommandlockouts
\usepackage[T1]{fontenc}
\usepackage{float}
\usepackage{amsfonts}
\usepackage{amsthm}
\usepackage{pifont}
\usepackage{xcolor}
\usepackage{soul}
\usepackage{url}

\newtheorem{proposition}{Proposition}

\usepackage{glossaries}
\usepackage{listings}
\usepackage{amsmath}
\usepackage{amssymb}
\usepackage{graphicx}
\usepackage{stix}
\usepackage{tikz}
\usepackage{etoolbox}
\usetikzlibrary{shapes,arrows}
\usepackage{verbatim}
\usepackage{arydshln}
\usepackage{subcaption}

\usepackage[colorinlistoftodos]{todonotes}

\usepackage{soul}

\renewcommand{\baselinestretch}{0.98}
\newcommand{\deleted}[1]{}

\begin{document}
\title{\bf 
SBlimp: Design, Model, and Translational Motion Control for a Swing-Blimp
%blimp
}
% \large{
% Design Parameters}
% }
\author{Jiawei Xu${}^{\dag}$, Diego S. D'Antonio${}^{\dag}$, Dominic J. Ammirato${}^{\dag}$, and David Salda\~{n}a${}^{\dag}$
\thanks{The authors gratefully acknowledge the support from Office of Naval Research Grant  N00014-23-1-2535}
\thanks{The authors acknowledge Alex Witt at Lehigh University for his help during the experiments.}
\thanks{${}^{\dag}$J. Xu, D. D'Antonio, D. Ammirato, and D. Salda\~{n}a are with the Autonomous and Intelligent Robotics Laboratory (AIRLab), Lehigh University, PA, USA: $\{$\texttt{jix519, diego.s.dantonio, dja223, saldana\}@lehigh.edu}
}}

\maketitle
% \begingroup\renewcommand\thefootnote{\textsection}
% \footnotetext{Equal contribution}
% \endgroup
%\tableofcontents 

\begin{abstract}
We present an aerial vehicle composed of a custom quadrotor with tilted rotors and a helium balloon, called SBlimp. 
We propose a novel control strategy that takes advantage of the natural stable attitude of the blimp to control translational motion.
Different from cascade controllers in the literature that controls attitude to achieve desired translational motion, our approach directly controls the linear velocity regardless of the heading orientation of the vehicle. As a result, the vehicle swings during the translational motion. We provide a planar analysis of the dynamic model, demonstrating stability for our controller. 
Our design is evaluated in numerical simulations with different physical factors and validated with experiments using a real-world prototype, showing that the SBlimp is able to achieve stable translation regardless of its orientation. 
% Experiments in an environment with obstacles demonstrate the blimp's ability to passively bounce and recover after collisions.%the resilience of the blimp against natural disturbance and collision with obstacles.

\end{abstract}

\section{Introduction}
%/include: (dynamics, controller, performance factors, implementation) 
Unmanned Aerial Vehicles (UAVs) have become a great interest for industry and academia. Some of the most popular vehicles are multi-rotor vehicles, which use rotors to generate thrust force to compensate for gravity and control its motion. Multi-rotor vehicles provide unparalleled design flexibility and aerial maneuverability. However, factors such as short flight duration and low payload capacity constrain their real-world applications.

To address the limitations of multi-rotor vehicles, we turn to a well-established design in the world of aerial vehicles and airships~\cite{M2021100741}. Non-rigid airships use large containers filled with lighter-than-air (LTA) gas, providing a natural buoyancy force that allows the vehicle to remain airborne with minimal energy expenditure to compensate for gravity~\cite{680971}. By integrating helium balloons and rotors using low-cost components, miniature robotic blimps can be created~\cite{680973}. Helium is a common gas in the universe, and when released into the atmosphere, it has no adverse environmental impact. Robotic blimps benefit from both the buoyancy provided by the LTA gas and the motion capabilities of a UAV, such as vertical takeoff and landing~\cite{Petrescu2017}. However, the large size of the balloon results in a low drag-to-lift ratio, which limits the agility of the vehicle.
Researchers have developed robotic blimps to complete tasks such as path following~\cite{9062572}, localization~\cite{muller2013efficient}, air surveillance~\cite{tan2012twin}, turbulence detection~\cite{arun2006atmos}, and formation control~\cite{9494636}. Moreover, blimps offer a high level of safety in cluttered environments, as the flexibility of the balloon allows it to absorb collisions with obstacles without incurring significant damage~\cite{doi:10.1177/1756829317705326,huang2019duckiefloat}.
% Vertical Takeoff and Landing (VTOL) platform~\cite{Petrescu2017}, the vehicle can achieve longer continuous flight time. 
% \st{With the same battery capacity, not being required to allocate a significant amount of energy to keep a blimp airborne means that it can sustain a longer flight duration and carry heavier payloads and equipment.} 
% {The property of enduring operation fits miniature blimps in a wide range of applications that do not require speed.} 
\begin{figure}[t]
    \centering
    \includegraphics[trim=13.cm 18cm 14cm 7cm,clip, width=0.8\linewidth]{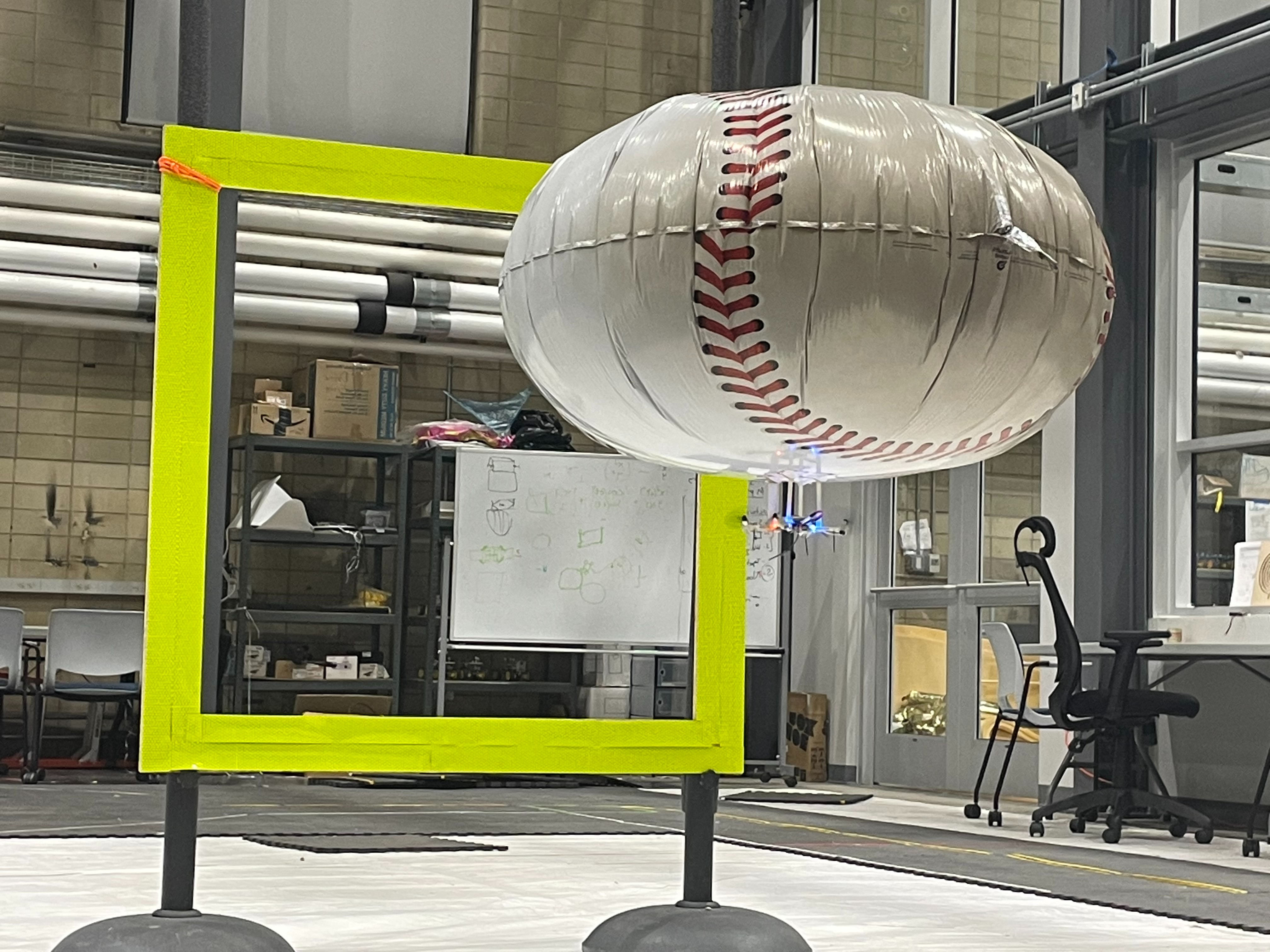}
    \caption{A flying SBlimp passing through a window. Narrated experiment videos can be found at \texttt{https://youtu.be/KN-64ZfBpFg}.}
    \label{fig:blimp}
    \vspace{-1.5em}
\end{figure}

Despite their wide range of applications, common blimp designs are associated with several disadvantages that require reconsideration. First, most existing blimp designs are underactuated and non-holonomic~\cite{liew2017recent,Wang2020}. Blimps with fixed rotor directions for forwarding and levitation are unable to translate backward without changing the heading direction~\cite{Qiuyang2018,sebbane2011lighter}, and those with differential models are unable to translate sideways~\cite{2021Stabilization}. Second, to achieve better actuation, the rotors typically reach out from the side of the balloon, which weakens their safety advantage.
Third, due to the need for sufficient buoyancy, blimps tend to be large, resulting in a high moment of inertia and vulnerability to environmental disturbances such as wind flow~\cite{ouerghi2022improved}. 
% The trade-off between motor power and movement speed complicates the control of blimps when fast motion is required. 
Uneven surface areas on the balloon lead to irregular swinging behaviors, deteriorating the quality of motion control~\cite{Qiuyang2018}. These factors present significant challenges to achieve effective control of blimps. 
% \qb{hi}
In light of these challenges, researchers have turned to multi-rotor UAV designs for inspiration, which have proven to offer higher motion control quality at the cost of added complexity. For example, in~\cite{yin2012}, the blimp used 4 servo motors to change the thrusting direction of the 4 rotors, and in~\cite{Burri2013}, the authors used 6 rotors to control the motion of the blimp. Both designs achieve full actuation. 
% For example, a blimp with two vertical rotors to adjust altitude and two front-facing horizontal rotors to steer is unable to translate backward without changing the heading direction~\cite{Qiuyang2018,sebbane2011lighter}.
% Blimps with a differential model using two actively tilting rotors are unable to translate sideways~\cite{2021Stabilization}. To achieve better actuation, the rotors typically reach out from the side of the balloon, which weakens their safety advantage.
% Additionally, to provide sufficient buoyancy, the size of a blimp is likely to be large, resulting in a high moment of inertia and the vulnerability to wind disturbances~\cite{ouerghi2022improved}. There is a trade-off between requiring high motor power for translation and movement speed. When the wind disturbance or the motion speed of the blimp is high, the uneven surface areas on the balloon lead to a swinging behavior that degrades the quality of the translational motion control~\cite{Qiuyang2018}. These factors further render the challenge in controlling a blimp precisely. Inspired by multi-rotor UAV designs, adding more rotors to a robotic blimp design contributes to higher motion control quality at the cost of weight, which can extend to precise holonomic motion control. In~\cite{yin2012}, the blimp used 4 servo motors to change the thrusting direction of the 4 rotors. In~\cite{Burri2013}, the authors used 6 rotors to control the motion of the blimp. Both designs achieve full actuation. 

In this paper, we present a novel and minimalistic design for a robotic blimp composed of a quadrotor with tilted propellers and an LTA balloon. We highlight that the quadrotor can fly independently~\cite{Xu2021} and the balloon can be considered as an extension module added to the quadrotor. We name the design \emph{SBlimp}. 
Notably, our SBlimp leverages its pendulum-like natural stability to achieve stable motion control solely through translational motion control despite the heading direction or the swinging behavior typically observed in traditional blimp designs. In general, our proposed design presents an innovative solution to the challenges posed by traditional blimp designs.

The main contribution of this paper is threefold. 
First, we propose a novel robotic blimp design composed of a quadrotor with tilted rotors rigidly attached to a balloon.
Second, we show the stability of the blimp controlled by the linear velocity controller in 2-D.
Third, we demonstrate a new motion behavior through numerical simulations and an actual prototype.
Compared to quadrotors, our SBlimp can remain airborne for more than ten times longer. Compared to other LTA airships, our SBlimp can translate without requiring attitude control. 

%
% \cite{Saldana2018}
% \cite{franchi2018full}
% \cite{modquadgripper}
% \cite{Qiuyang2018}

% related works
% \cite{Burri2013}

\section{Design}
\begin{figure}[t]
    \centering
    \includegraphics[trim=0cm 2cm 0cm 0cm, clip, width=0.8\linewidth]{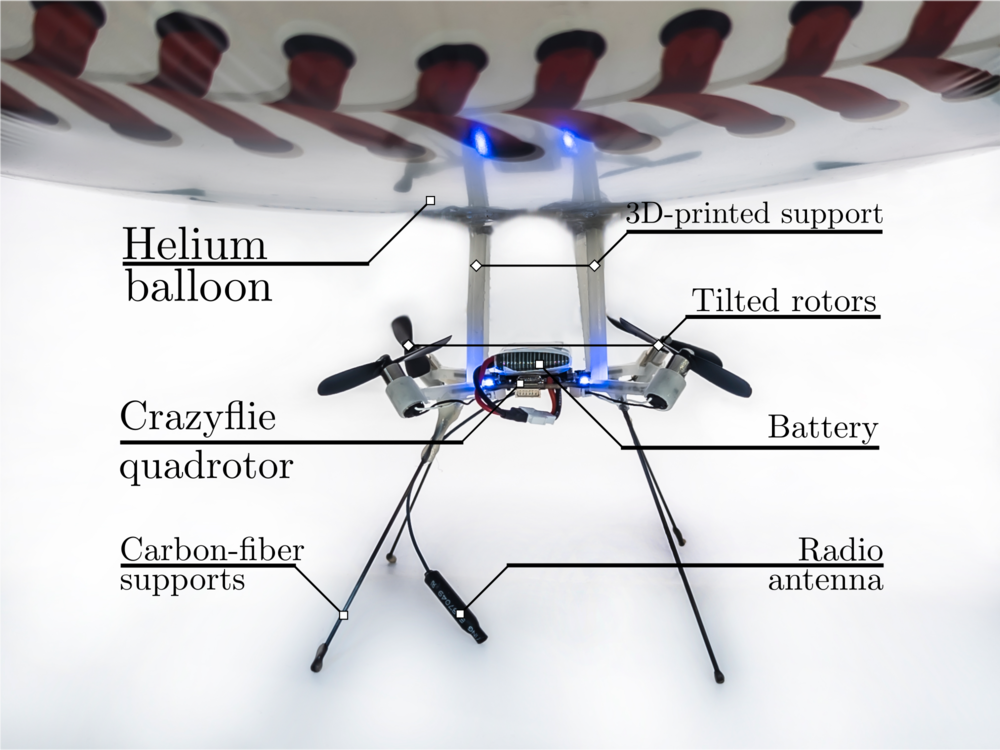}
    \caption{Components of a SBlimp.}
    \vspace{-1em}
    \label{fig:blimpcomp}
\end{figure}
\label{sec:design}
The design of the SBlimp is minimalistic, composed of two main parts (see Fig.~\ref{fig:blimpcomp}):
\paragraph{Quadrotor} The vehicle is propelled by the quadrotor Crazyflie 2.1, an open-source software and hardware platform. 
%As a micro UAV, the Crazyflie is perfect for our experimental purposes as larger and more massive quadrotor platforms would require much larger blimps to support them.
We modify the Crazyflie 2.1 design with 3D printed motor mounts that accommodate our tilted rotor arrangement. The design is inspired by our previous vehicle, called $T$-module~\cite{Xu2021}. 
% cflib python library and the crazyradio communication protocol~\cite{crazyflieROS} to send commands from the base station to the UAV.
    
\paragraph{Helium Balloon} The balloon is made of Mylar and has the shape of an ellipsoid. The total volume of the ellipsoid is $0.125\; m^3$ and is filled with industry-grade helium with a concentration of $99\%$ helium.
%From our measurements, the lift force of the blimp alone is equal to $0.66\; N$\david{In grams? }. 
%The symmetric ellipsoidal balloon avoids discrepancies between the moment of inertia in roll and pitch.
    
The two parts are connected using a 3D-printed support with Formlabs Durable material. The quadrotor is centered below the intersection of the two major axes of the ellipsoid. 
The low center of mass (COM) offers a natural stability~\cite{7106503} that is discussed in Sec.~\ref{sec:control}. The distance between the center of lift (COL) of the blimp and the COM of the quadrotor, $L_b$, is measured as $0.35\; m$. With the $1$-cell ($3.7\; V$) $750\; mAh$ battery, the SBlimp weighs $60$ grams (without helium). When the balloon is filled with helium, the negative buoyancy of the vehicle is $5\; g$.

% We construct the prototype quadrotor-blimp based on the Crazyflie 2.1 platform, which is an open-source software and hardware development platform by Bitcraze. We modify the base design using 3D printed parts accommodating our tilted rotor scheme and allow for attachment to the blimp~\cite{Xu2021}. The design factor $\eta = \frac{\pi}{4}$. We use the cflib python library and the crazyradio communication protocol~\cite{} to send commands from the base station to the blimp. For localization, we use the motion capture system (Optitrack) operating at 120 Hz. The crazyflie also measures its angular velocities using an onboard IMU sensor, and this data is merged with the external pose data through the crazyflie's own implementation of the Extended Kalman filter~\cite{mueller2015fusing}. The balloon attatched to the crazyflie is made of mylar, contains consumer grade helium of about $0.125\; m^3$ volume, which is listed as having no less than a concentration of $80\%$ helium. From our measurements, $L_b = 0.35\; m$, $f_b = 0.32\; N$. With the stock $1$-cell ($3.7\; V$) $370\; mAh$ battery, the quadrotor-blimp weighs $35$ grams without the helium.

\section{Model}
\begin{figure}[t]
    \centering
    \includegraphics[width=0.8\linewidth]{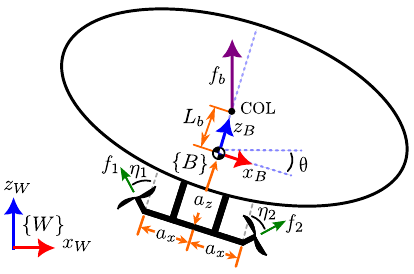}
    \caption{The model of the SBlimp in 2-D. $\{W\}$ and $\{B\}$ represent the world reference and the SBlimp frame, respectively. $L_b$ is the distance between the COL (marked with a solid dot) and the COM (marked with a dowel pin symbol). The pitch~$\theta$ describes the tilting of the blimp.
    }    
    \label{fig:blimpfigure}
    \vspace{-2em}
\end{figure}

Our objective is to design a control strategy that enables the SBlimp vehicle to move in any direction regardless of its orientation.
This paper focuses on analyzing the planar configuration of the vehicle. 
We define the world reference frame as a fixed frame, with $z$-axis pointing upward, denoted by $\{W\}$. The blimp has a body frame $\{B\}$ with the origin at the COM. The $x$-axis points toward the front of the blimp and the $z$-axis points upwards, as illustrated in Fig.~\ref{fig:blimpfigure}. 

The helium balloon provides a buoyancy force $f_b > 0$ always pointing in the direction of $z_W$, and the quadrotor is equipped with tilted propellers. 
% The $z$-axis of $\{M\}$ points vertically up, and i$x$-axis is oriented $90$-degrees counterclockwise from the $z$-axis.
In the planar case, the quadrotor has two rotors, tilted at angles of $\eta_1$ and $\eta_2$ with respect to the $z$-axis of $\{B\}$, satisfying $\eta_1 = -\eta_2$.
The positions of the rotors in $\{B\}$ are $\boldsymbol{p}_1=\left[a_x, a_z\right]^\top$ and $\boldsymbol{p}_2=\left[-a_x, a_z\right]^\top$. Letting $\eta = \eta_2$, we characterize the blimp with a design factor $-\frac{\pi}{2} < \eta < \frac{\pi}{2}$. % The COM of the drone is at the middle point of the two rotors. By aligning the COM of the entire blimp on the $z$-axis of $\{M\}$, 
% The vertical distance between the COM of the base and the blimp is $\vert p_z\vert$.
We put the rotors below the COM of the blimp so that $a_z < 0$. 
%In Sec.~\ref{sec:control}, we will show how this design allows the blimp to achieve stability. 
The COL of the balloon lies on the positive $z$-axis of~$\{B\}$, of which the distance to the COM of the blimp is $L_b$. 
The position of the vehicle in~$\{W\}$ is denoted by $\boldsymbol{r} = \left[x, z\right]^\top\in\mathbb{R}^2$. 
Vectors $\boldsymbol{v = \dot{r}}$ and $\boldsymbol{\dot{v} = \ddot{r}}$ denote linear velocity and acceleration, respectively. 
%The orientation of the vehicle is denoted by $\theta$.
The orientation of the blimp, ${}^W\!\!\boldsymbol{R}\!_B$, is described by the rotation from $\{W\}$ to $\{B\}$. Denoting the tilting angle by the pitch $\theta$, we have ${}^W\!\!\boldsymbol{R}\!_B = \text{Rot}(\theta)$ where the 2-D rotation matrix operator $\text{Rot}(\alpha) = \begin{bmatrix}
\cos{\alpha} & -\sin{\alpha}\\
\sin{\alpha} & \cos{\alpha}
\end{bmatrix}$. 
Due to the symmetric orthogonal dynamics of the blimp in the $xz$- and $yz$-plane, the analysis based on the planar model can be extrapolated to the 3-D case. 
% For ease of analysis, we focus on the planar configuration.

The planar SBlimp uses two rotors to generate thrust.
We denote the thrust force by $f_i$, for $i\in\{1,2\}$. Then the thrust vectors in $\{B\}$ are
$\boldsymbol{f}_{i} = [-f_{i}\sin{\eta_i}, f_{i}\cos{\eta_i}]^\top.$
% $
% \boldsymbol{f}_{i}=f_{i}\:  {{}^{B}\!\boldsymbol{R}_{i} \boldsymbol{\hat z}}
% ,
% $
% where ${}^{B}\!\boldsymbol{R}_{i} = \text{Rot}(\eta_i)$ is the 2-D rotation matrix that represents the direction of the thrust forces, and the unit vector $\boldsymbol{\hat z} = \left[0, 1\right]^\top$. This notation means we can obtain the direction of thrust $f_i$ in $\{B\}$ by rotating $\boldsymbol{\hat z}$ for an $\eta_i$ angle. 
The positions of the rotors $\boldsymbol{p}_i$ are the arms of the thrust forces from the COM of the blimp. Therefore, each rotor also applies a torque $\tau_{i} = \boldsymbol{p}_{i}\times\boldsymbol{f}_{i}
% = \Vert\boldsymbol{p}_{i}\Vert {f}_{i} {\sin{\eta_i}}
$ 
% \david{$\tau_1 = \boldsymbol{p}_{i}\cdot\boldsymbol{f}_{i}$.}
on the blimp in pitch. Thus, the total force~$\boldsymbol{f}$ and torque~$\tau$ in~$\{B\}$ generated by the rotors $
    \boldsymbol{f} = \boldsymbol{f}_{1} + \boldsymbol{f}_{2}, \text{ and }
    \tau = \tau_{1} + \tau_{2}.$

Denoting $\boldsymbol{u} = \left[f_{1}, f_{2}\right]^{\top}$ as the input vector, we can express the force and torque applied by the rotors on the blimp in the matrix form,
$
    \boldsymbol{f} = \boldsymbol{A_f}\boldsymbol{u}, \text{ and }
    \tau = \boldsymbol{A}_\tau\boldsymbol{u}
$
where %the \emph{allocation matrices} are
\begin{equation}
\begin{aligned}
    &\boldsymbol{A_f}%(\eta)
    &=&
    \begin{bmatrix}
        \sin{\eta} & -\sin{\eta}\\
        \cos{\eta} & \cos{\eta}
    \end{bmatrix}, \text{ and }\\
    &\boldsymbol{A}_\tau%(\eta, a_x, a_z) 
    &=& 
    \begin{bmatrix}
        a_z\!\sin{\eta}\!-\!a_x\!\cos{\eta} &
        a_x\!\cos{\eta}\!-\!a_z\!\sin{\eta}
    \end{bmatrix}
\end{aligned}
\label{eq:Acomponents}
\end{equation}
are matrices that map the input forces into the total force and torque in $\{B\}$.

Due to the large volume of the balloon, the blimp experiences significant air drag on its surface. According to~\cite{batchelor1967introduction}, the air drag depends on the Reynold number. Under low speed, the low Reynold number results in an air drag approximately proportional to the velocity~\cite{lissaman1983low}. Therefore, we incorporate the air drag by modeling the dissipative force as first-order damping terms for translation $\boldsymbol{D} = \text{diag}\left(\left[d_x, d_z\right]^\top\right)$ and rotation $d_\tau>0$, where $d_x, d_z>0$ are the air drag coefficients. 
The LTA helium balloon generates a buoyancy force $f_b$ at a distance $L_b$ from the COM, generating a torque,
% $\boldsymbol{\tau}_g=
% \left[
% 0, 0, 0
% \right]^\top$, 
\begin{equation}
    \tau_b=\left[0, L_b\right]^\top\times\left({}^B\!\boldsymbol{R}_{\!W}\!\!\left[0, f_b\right]^\top\right) =   -f_bL_b\sin{\theta},
    \label{eq:taub}
\end{equation}
% Note that the yaw angle of the blimp only rotates the arm of the helium force around the $z$-axis. Therefore, $\boldsymbol{\tau}_b = \left(\text{rot}_y(\phi)\text{rot}_x(\theta)\left[
% 0, 0, L_b\right]^\top\right)\times\left[0, 0, f_b\right]^\top$ only depends on the roll and pitch of the blimp. 
% Furthermore, a normal quadrotor whose body frame has its origin at its center of mass (COM). Its gravity generates zero torque around its COM. 
% when ${}^W\!\!\boldsymbol{R}_B\neq\boldsymbol{I}_3$. 
where ${}^B\!\boldsymbol{R}_{\!W} = {}^W\!\!\boldsymbol{R}\!_B^\top$. For any 
$\theta\neq0$, the torque $\tau_b\neq0$.

We describe the dynamics of the blimp using Newton-Euler equations,
\begin{align}
    m\boldsymbol{\dot v} &= -\boldsymbol{D}{\boldsymbol{v}} + (f_b - mg)\boldsymbol{\hat z} + {}^W\!\!\boldsymbol{R}\!_B\boldsymbol{A_fu},\label{eq:Newton}\\
    J_\theta\ddot{\theta} &= -d_\tau\dot{\theta} + \tau_b + \boldsymbol{A}_\tau\boldsymbol{u},
    \label{eq:Euler}
\end{align}
where the $m$ is the mass of the blimp, $J_\theta$ is the moment of inertia, $g$ is the gravitational acceleration, and $f_b$ is the buoyancy force provided by the balloon. 
%Since we set the origin of $\{B\}$ at the COM of the blimp, the gravity applied on the total mass of the blimp $m$ contributes zero torque on the blimp. Since 
% \begin{problem}[Control the blimp]
% Given a desired position $\boldsymbol{r}^d$, find the input $\boldsymbol{u}$ such that drives the blimp position $\boldsymbol{r}$ to stabilize at $\boldsymbol{r}^d$.
% \end{problem}
Based on our SBlimp model, the vehicle naturally tries to stay horizontal. We focus on controlling its translational velocity, \textit{i.e.}, given a desired velocity $\boldsymbol{v}^d$, we find an input $\boldsymbol{u}$ that drives the SBlimp to a velocity of $\boldsymbol{v}^d$ without controlling its attitude.

\section{Control Design}
\label{sec:control}

\begin{figure}
    \centering
    \includegraphics[trim=1cm 0cm 0.8cm 0cm, clip, width=.9\linewidth]{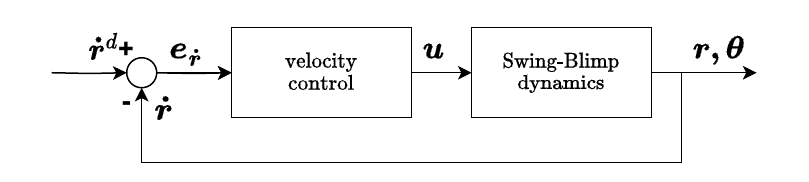}
    \caption{The flow diagram of the SBlimp controller, where the feedback linearization involves only the velocity.}
    \label{fig:controldiag}
    \vspace{-1em}
\end{figure}

We design a velocity controller for the SBlimp using its inherent rotational stability.
An overview of our controller is illustrated in Fig.~\ref{fig:controldiag}. The objective of our controller is to drive the blimp to a desired velocity, $\boldsymbol{\dot{r}}^d$. 

When the blimp is not actuated, \textit{i.e.}, $\boldsymbol{u}=\boldsymbol{0}$, the system in~\eqref{eq:Euler} is reduced to $J_\theta\ddot{\theta} = -d_\tau\dot{\theta} + \tau_b$. Substituting the buoyancy torque $\tau_b$ from \eqref{eq:taub}, the differential equation describes a behavior that is similar to a damped pendulum model~\cite{baker2008pendulum}, which is stable for any initial $\theta$. 

\subsection{Feedback Linearization}
We use the quadrotor to control the translational motion of the blimp.
% The allocation matrices of a traditional quadrotor has the first row containing only zeros because their rotors can only generate forces along the $z$-axis in its body frame, rendering its matrix $\boldsymbol{A_f}$ rank-deficient. Such property does not change if we attach it to a balloon. In contrast, 
With tilted rotors, \textit{i.e.}, $0<\eta<\frac{\pi}{2}, {}^{B}\!\boldsymbol{R}_{i}\neq\boldsymbol{I}$, the allocation matrix $\boldsymbol{A_f}$ is full-rank and invertible, allowing the vehicle to generate forces in any direction within the rotor constraints. We will use this property to achieve the control objective. 
Since the rotation matrix ${}^W\!\!\boldsymbol{R}\!_B$ is observable and invertible, we apply feedback linearization on~\ref{eq:Newton}. Ignoring the damping force, we can choose the input
\begin{equation}
    \boldsymbol{u} = m\boldsymbol{A_f}^{\!\!\!-1}\:{}^B\!\boldsymbol{R}_{\!W}\left(\left(g-\frac{f_{b}}{m}\right)\boldsymbol{\hat z} + \boldsymbol{w}\right),
    \label{eq:u}
\end{equation}
where ${}^B\!\boldsymbol{R}_{\!W} = {}^W\!\!\boldsymbol{R}_{\!B}^{-1} = {}^W\!\!\boldsymbol{R}_{\!B}^{\top}$, and $\boldsymbol{w} = \boldsymbol{\dot v}$ is the additional input, which shows that the system is linearized via static feedback. We apply a proportional control on the velocity,
\begin{equation}
    \boldsymbol{w} = 
    % \boldsymbol{K_r}\boldsymbol{e_r} + 
    \boldsymbol{K_{v}}\left(\boldsymbol{v}^d - \boldsymbol{v}\right),
    \label{eq:w}
\end{equation}
where $\boldsymbol{K_{v}} = \text{Diag}\left(\left[k_{\dot x}, k_{\dot z}\right]^\top\right)$ is the positive-definite gain matrix. 

\subsection{Stability analysis}
We first show the stability in velocity tracking of the blimp with our controller. Inserting \eqref{eq:u} and \eqref{eq:w} into \eqref{eq:Newton}, we obtain the closed-loop velocity dynamics, 
\begin{equation}
    m\boldsymbol{\dot v} = \boldsymbol{K_{v}}\left(\boldsymbol{v}^d - \boldsymbol{v}\right) - \boldsymbol{D}\boldsymbol{v}.
    \label{eq:2DposDynamics}
\end{equation}
which takes the form of a first-order linear differential equation. Solving the equation using the integrating factor method for $\boldsymbol{v}$ gives the linear velocity response of the controlled SBlimp,

\begin{equation}
    \boldsymbol{v} = \begin{bmatrix}
        \frac{k_{\dot x}}{k_{\dot x}+d_{x}} & 0\\
        0 & \frac{k_{\dot z}}{k_{\dot z}+d_{z}}
    \end{bmatrix}\boldsymbol{v}^d + \begin{bmatrix}
        e^{-\frac{k_{\dot x}+d_{x}}{m}t} & 0\\
        0 & e^{-\frac{k_{\dot z}+d_{z}}{m}t}
    \end{bmatrix}\boldsymbol{v}_0,
    \label{eq:velsol}
\end{equation}
where $\boldsymbol{v}_{0}$ is the initial velocity of the blimp, and $t$ is the time. With sufficiently large gains $k_{\dot x}\gg d_{x}, k_{\dot z}\gg d_{z}$ the velocity in $xz$-plane exponentially converges to the desired value as $t\!\rightarrow\!\infty$ up to an arbitrary precision depending on the gains, showing the exponential stability of the closed-loop velocity dynamics of the SBlimp.

Even though the linear and angular accelerations are coupled, we can show that the system is stable around $\theta=0$.
% We show that the system is stable around $\theta=0$. 

\begin{proposition}[The pitch angle is asymptotically stable around $\theta=0$]
    If the additional input $\boldsymbol{w} = 
    \boldsymbol{K_{v}}\left(\boldsymbol{v^d - v}\right)$ only takes feedback on velocity, the pitch $\theta$ in \eqref{eq:Euler} can still converge to $0$ asymptotically.
\end{proposition}
\begin{proof}
Inserting \eqref{eq:u} and \eqref{eq:w} into \eqref{eq:Euler}, we obtain the closed-loop angular dynamics, 
\begin{equation}
    J_\theta\ddot{\theta} = -d_\tau\dot{\theta} + \tau_b + \boldsymbol{A}_\tau\boldsymbol{A_f}^{\!\!\!-1}\:{}^B\!\boldsymbol{R}_{\!W}\left(\left(mg-f_{b}\right)\boldsymbol{\hat z} + m\boldsymbol{w}\right).
    \label{eq:tauClosedLoop}
\end{equation}
Notably, the allocation matrices $\boldsymbol{A}_\tau$ and $\boldsymbol{A_f}$ are not linearly independent for a SBlimp. The coupling relationship is described as $\boldsymbol{A}_\tau = \left[c\; 0\right]\boldsymbol{A_f}$, where the coefficient $c = a_z-\frac{a_x}{\tan{\eta}}$ depends on $\eta$, $a_x$, and $a_z$. Therefore,~\eqref{eq:tauClosedLoop} is simplified to
\begin{equation}
    J_\theta\ddot{\theta} = c\begin{bmatrix}
        \cos{\theta}\\
        \sin{\theta}
    \end{bmatrix}^\top\left(\left(mg-f_{b}\right)\boldsymbol{\hat z} + m\boldsymbol{w}\right) - (d_\tau\dot{\theta}+f_bL_b\sin{\theta}).
    \label{eq:2DangDynamics}
\end{equation}

We linearize the angular dynamics around $\theta = 0$ by taking the Taylor expansion of $\cos{\theta}$ and $\sin{\theta}$. As $\theta\rightarrow0$, $\cos{\theta} = \sum_n^\infty(-1)^n\frac{\theta^{2n}}{(2n)!}\rightarrow1$, and $\sin{\theta} = \sum_n^\infty(-1)^n\frac{\theta^{2n + 1}}{(2n + 1)!}\rightarrow\theta$, considering $n = 0$ and discarding all higher order terms. Therefore,~\eqref{eq:2DangDynamics} becomes\begin{equation}
    J_\theta\ddot{\theta} = 
    c\begin{bmatrix}
        1 & \theta
    \end{bmatrix} \boldsymbol{K_{v}}\left(\boldsymbol{v^d - v}\right)
    - (d_\tau\dot{\theta}+f_bL_b\theta),
    \label{eq:2DangDynamicsLinear}
\end{equation}
which depends on the velocity tracking error of the blimp. By replacing $\boldsymbol{v}$ with the general solution of the closed-loop velocity response \eqref{eq:velsol}, we obtain the second-order differential equation of $\theta$. The general solution to the non-homogeneous differential equation is in the form of the multiplication of Bessel functions~\cite{abramowitz1964handbook} and negative-exponent exponential functions of $t$. The Bessel function highlights the vibration behavior of $\theta$ near $\theta = 0$, which results in the swinging of our SBlimp. The exponential components describe the dissipated kinetic energy as $t\rightarrow\infty$ regardless of the initial value of $\theta$, which shows the asymptotic stability of the pitch $\theta$ around its natural equilibrium~$\theta=0$.
\end{proof}

Numerical approximations of the non-linearized angular dynamics as described in~\eqref{eq:2DangDynamics} also show the convergence of $\theta$ to $0$ as $t\rightarrow\infty$, which further demonstrates the damped pendulum-like stability of the pitch angle.

\section{Evaluation}
To evaluate the performance and limits of our design and control, we conduct experiments in simulations and with physical robots\footnote{The source code of our simulation and experiments can be found at https://github.com/swarmslab/OpenBlimp}.
% To assess the efficacy of the QBlimp's design and control,  This simulator enables us to conduct simulated experiments and evaluate the performance of the blimp with physical parameters.
We build a prototype SBlimp as described in Section~\ref{sec:design} and test its ability to follow increasingly complex trajectories, validating the effectiveness of the controller and demonstrating the blimp's significantly elongated hovering time. In simulation, we focus on metrics of velocity error and pitch swinging angle, while in experiments, we use position error as our primary evaluation metric.

% Our experiments with obstacles show that the design is collision-tolerant.} 

\subsection{Simulation}
We implement a 2-D numerical simulator using Python. We leverage the convenience of manipulating various physical factors that influence the performance of the blimp, which would otherwise be time-consuming to investigate on a physical prototype, including the distance between the COL and COM $L_b$, the mass $m$, and the target speed. In each set of tests, the blimp follows a circular path of $1$-meter radius for $100$ seconds while recording velocity and angular errors. The maximum and minimum rotor power of each of the two rotors are set at $0.15\; N$ and $0$ respectively. We refer to a rotor as ``saturated'' when the required force output from the controller goes beyond these limits.

\subsubsection{Distance $L_b$} We evaluate the effect of $L_b$ in the performance, the distance between the COM and the COL of the blimp. During the early prototyping stage, the test flights indicated that $L_b$ affects \textit{i)} whether the rotors can obtain enough airflow to generate the target thrust; and \textit{ii)} the frequency and amplitude of the pendulum behavior of the SBlimp. Although the simulator is not capable of simulating the aerodynamics of the system, we consider the pendulum behavior a key factor to the stability of the SBlimp. We increase the length $L_b$ from $0.01\; m$ to $1.0\; m$ subsequently and keep all other factors identical throughout the tests, including the target circular trajectory at a speed of $0.1\; m/s$. The mass of the blimp is $m = 0.06\; kg$, and the buoyancy provided by the balloon is $f_b = 0.55\; N$. 
% Since larger $L_b$ indicates a longer support between the quadrotor and the balloon, an additional weight of the support $m_s = 0.05L_b$ contributes to the overall weight of the blimp $m = m_0 + m_s$. 
Our simulations reveal that larger values of $L_b$ result in lower maximum and average angular errors, while the maximum velocity error remains below $0.01\; m/s$. However, beyond $L_b = 0.3\; m$, the angular error reduction becomes marginal, as the maximum angular error is below $0.02^\circ$. The result suggests that we are able to reduce the swinging behavior by increasing the length of the support. However, such design benefits may be offset by practical concerns such as reduced support rigidity and increased weight and inertia. We, therefore, conclude that $L_b = 0.3\; m$ is an appropriate choice for balancing between performance and practical considerations in the following simulations. 
% As $L_b$ goes higher than $2.0\; m$, the rotors start to saturate trying to hover and the positional errors increase. As a result, we keep $L_b = 0.3\; m$ for the following simulations. 

\begin{figure}[t]
    \centering
    \includegraphics[trim=0.2cm 0.8cm 0cm 1cm, clip,width=0.5\textwidth]{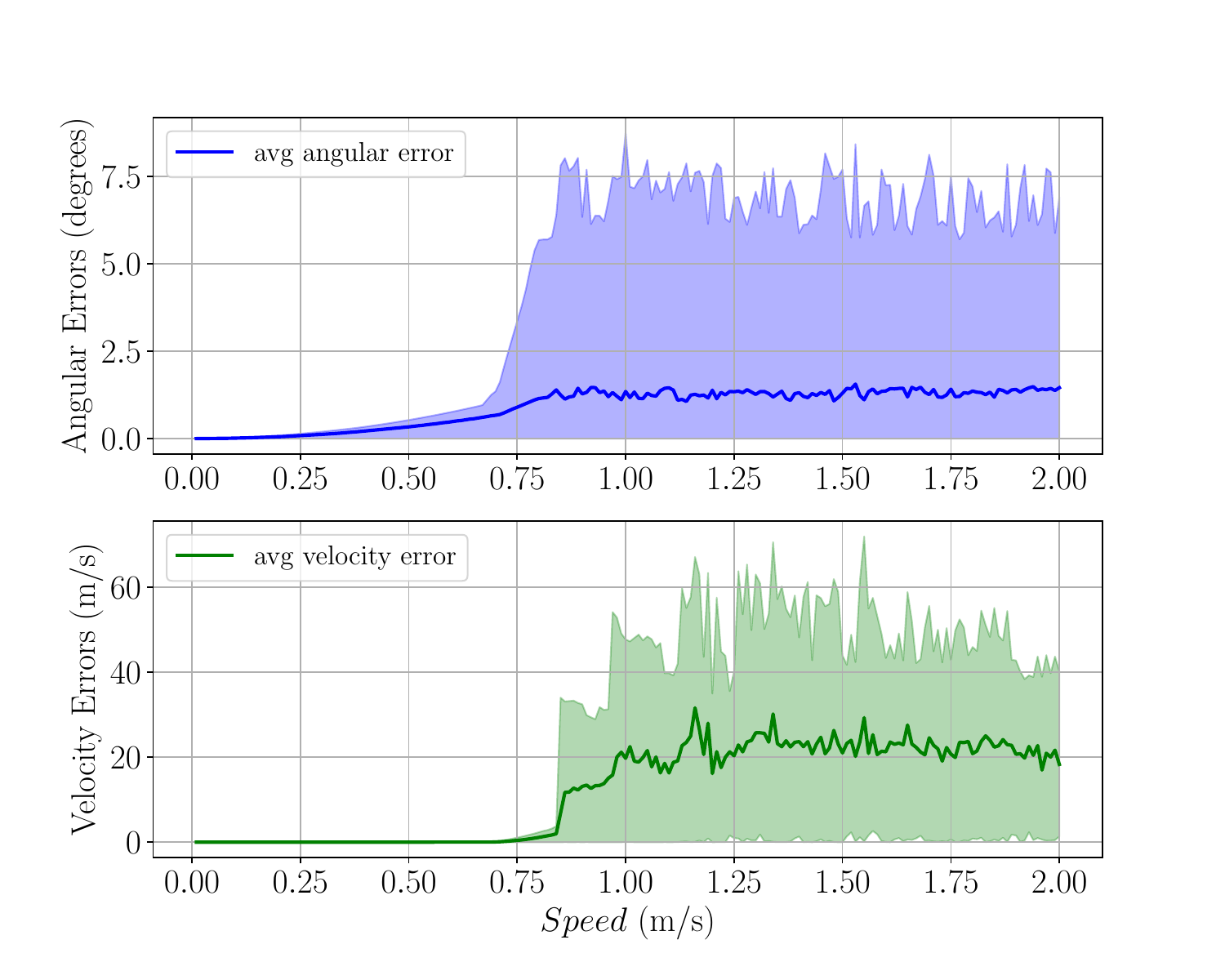}
    \caption{The average tracking errors of the SBlimp as the target speed increases with their maximum and minimum range. The maximum and average errors increase linearly as the target speed increases until the rotor saturation at the target speed of $0.6\; m/s$.}
    \vspace{-1em}
    \label{fig:simspeed}
\end{figure}

\subsubsection{Mass} We test how the increased mass of the quadrotor $m$ can affect the performance as if the SBlimp is carrying heavier equipment and batteries. We increase $m$ from $0.05\; kg$ to $0.1\; kg$ subsequently and keep other factors constant, \textit{i.e.}, $L_b = 0.3\; m$, $f_b = 0.55\; N$, and the target speed of $0.1\; m/s$. The simulations show that when the rotors are not saturated, the angular and velocity errors remain close to $0$ as $m$ increases. An increased mass only results in increased rotor force. When the gravity of the blimp is less than the buoyancy or the maximum thrust combined with the buoyancy cannot compensate for the gravity, the blimp loses control over its height.

\subsubsection{Speed} We evaluate the effect of varying speeds on the performance. Because of its low lift-to-drag ratio, the blimp experiences an increasing difficulty following higher velocity commands. 
% Consequently, the blimp should experience better performance at lower speeds. 
Therefore, we would like to know the theoretical limits of the speed for the blimp to maintain a high tracking quality.
We test target speeds from $0.01\; m/s$ to $2.0\; m/s$ with an increment of $0.01\; m/s$. As expected, the results of this simulation show that the maximum and average error increases almost linearly against the target speed until the rotor saturation, where the target speed is $0.6\; m/s$. As shown in Fig.~\ref{fig:simspeed}, when the target speed reaches $0.6\; m/s$, the swinging in pitch worsens dramatically. As the target speed exceeds $0.8\; m/s$, the controller loses stability in velocity tracking.

\subsection{Experiments with a prototype}
We build a prototype of the SBlimp with $L_b = 0.35\; m$, $m = 0.06\; kg$, and $f_b = 0.55\; N$.
The localization of the vehicle is realized using a motion capture system (Optitrack) operating at $120$ Hz. The angular velocity and linear acceleration are measured directly from Crazyflie 2.1's integrated IMU sensor. These two data sources are merged through Crazyflie's onboard Extended Kalman filter~\cite{mueller2015fusing}. We use the crazyswarm framework~\cite{crazyswarm} to communicate the desired linear velocity and yaw angle to the robot via the crazyradio communication protocol~\cite{crazyflieROS}. During the experiments, we send desired positions and velocities along a trajectory to the Crazyflie. With the prototype, we evaluate the errors in position when following trajectories of different target speeds and complexities.

\begin{figure}[t]
    \centering
    % \begin{subfigure}{1.\linewidth}
    % \includegraphics[trim=0.2cm 0cm 0cm 0cm, clip, width=1.\textwidth]{figures/circle_xy.pdf}
    % \caption{Circular Trajectory with low translational error in $xy$-plane.\david{what is this figure trying to show?}}
    % \label{fig:circle_a}
    % \end{subfigure}
    %%
    % \begin{subfigure}{1.\linewidth}
    \includegraphics[trim=0.2cm 0cm 0cm 0cm, clip, width=0.9\linewidth]{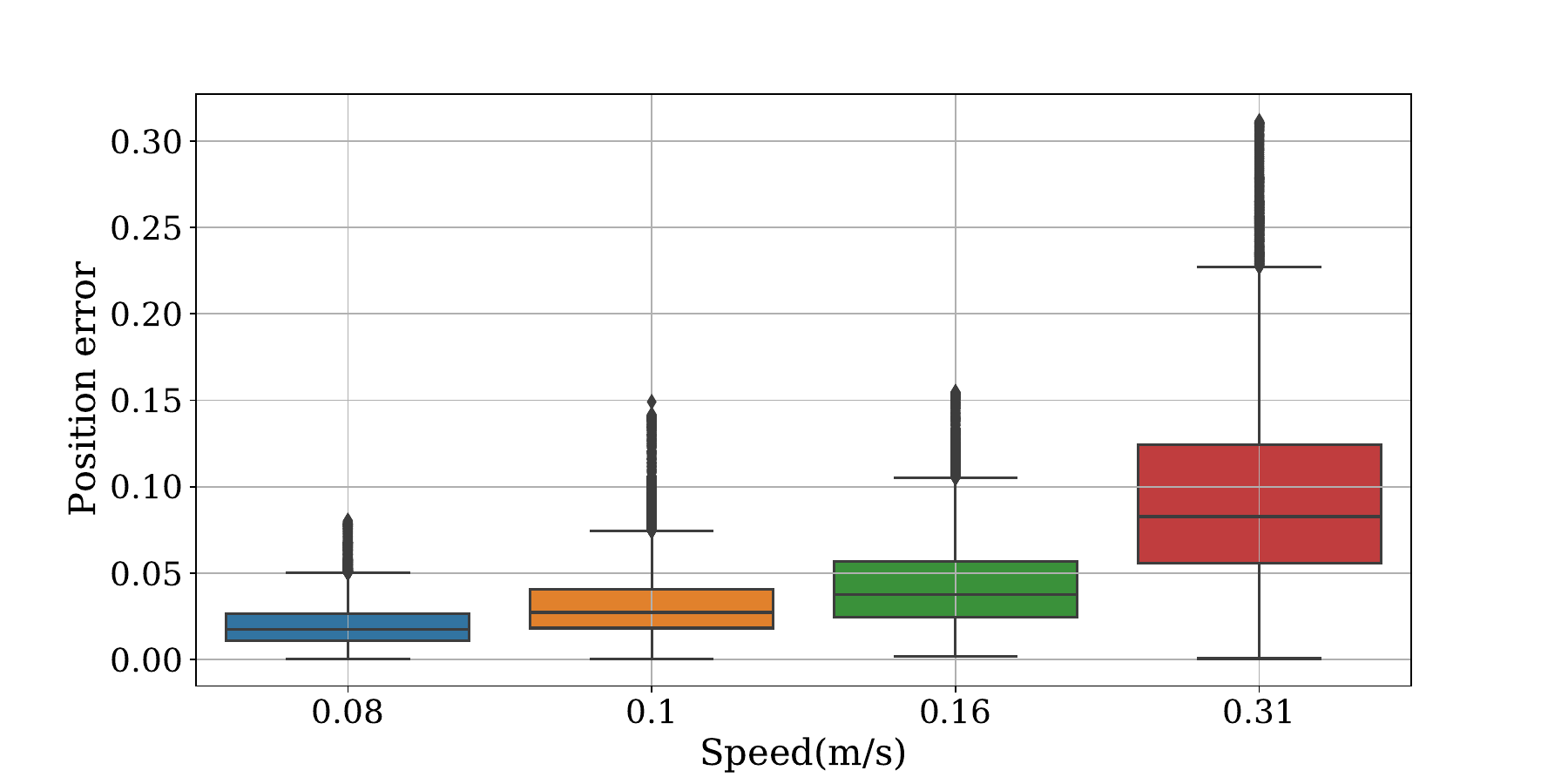}
    % \caption{A box whisker plot showing the magnitude of the position errors against increasing target velocities.}
    % \label{fig:circle_b}
    % \end{subfigure}
    \caption{The position error magnitudes of the SBlimp following the circular trajectory at different speeds. 
    The Average and maximum errors increase as the speed increases.}
    % The magnitude of the translational error remains less than $0.05$ m and achieves an average of $0.02$ m at a speed of $0.1$ m/s.}
    \label{fig:circle} 
    \vspace{-1em}
\end{figure}

\subsubsection{Experiment - Hovering}
{
We evaluate the blimp's ability to hover at a fixed position. 
% This capability is critical to the ongoing development of the blimp prototype, as it demonstrates its stability and convergence rate.
% The blimp was commanded to hover at a fixed point in ${W}$ at the origin and $1 \ m$ above the ground, and the distance between its initial and hovering positions was $2 \ m$.
%
% The blimp's position error in the $x$-, $y$-, and $z$-axes converged to zero in $25$ s, and 
The average magnitude of the position error during hovering was $0.01\; m$. 
Our experiments showed that \textbf{the robot can hover for 1 hour and 15 minutes}, which is more than ten times longer than the airborne duration achieved by the quadrotor alone.
%
% This experiment's results demonstrate the blimp's potential for prolonged stable hovering, making it a promising platform for various applications, such as aerial surveillance and monitoring.
}

\subsubsection{Experiment - Tracking a circular trajectory}
{
To demonstrate the independence of blimp translation on its heading direction, we conduct experiments in which the SBlimp follows a circular trajectory in the $xy$-plane without rotation. The trajectory is a circle with a radius of $1\; m$ and a fixed yaw of $0$ degrees, described by linear velocities $\dot x = -\sin{\frac{2\pi}{v}t}$ and $\dot y = \cos{\frac{2\pi}{v}t}$, where $v$ is the constant target speed and $t$ is the time from the beginning of the experiment.
We change $v$ between experiments to evaluate the speed limit of our system. 
The results are shown in Fig.~\ref{fig:circle}. 
%
% Each experiment lasts $30$ minutes, during which the blimp tries to follow a desired position and velocity along the trajectory.
Since the actuation in the $x$- and $y$- coordinates are coupled with the pitch and roll, the vehicle exhibits swinging behaviors. We observe that a higher target speed result in larger position errors. At $v = 0.1\; m/s$, the average error is $0.05\; m$, which is higher than hovering. At $v = 0.31\; m/s$, the average error is around $0.1\; m$. The increased error indicates that the rotors experience saturation when trying to achieve the required speed during swinging. 
Overall, our experiment successfully demonstrates that the SBlimp can control translational motion despite the swinging behavior.
}

\begin{figure}[t]
    \centering
    \begin{subfigure}{0.75\linewidth}
    \includegraphics[trim=7.0cm 2.cm 5cm 4cm, clip, width=\linewidth]{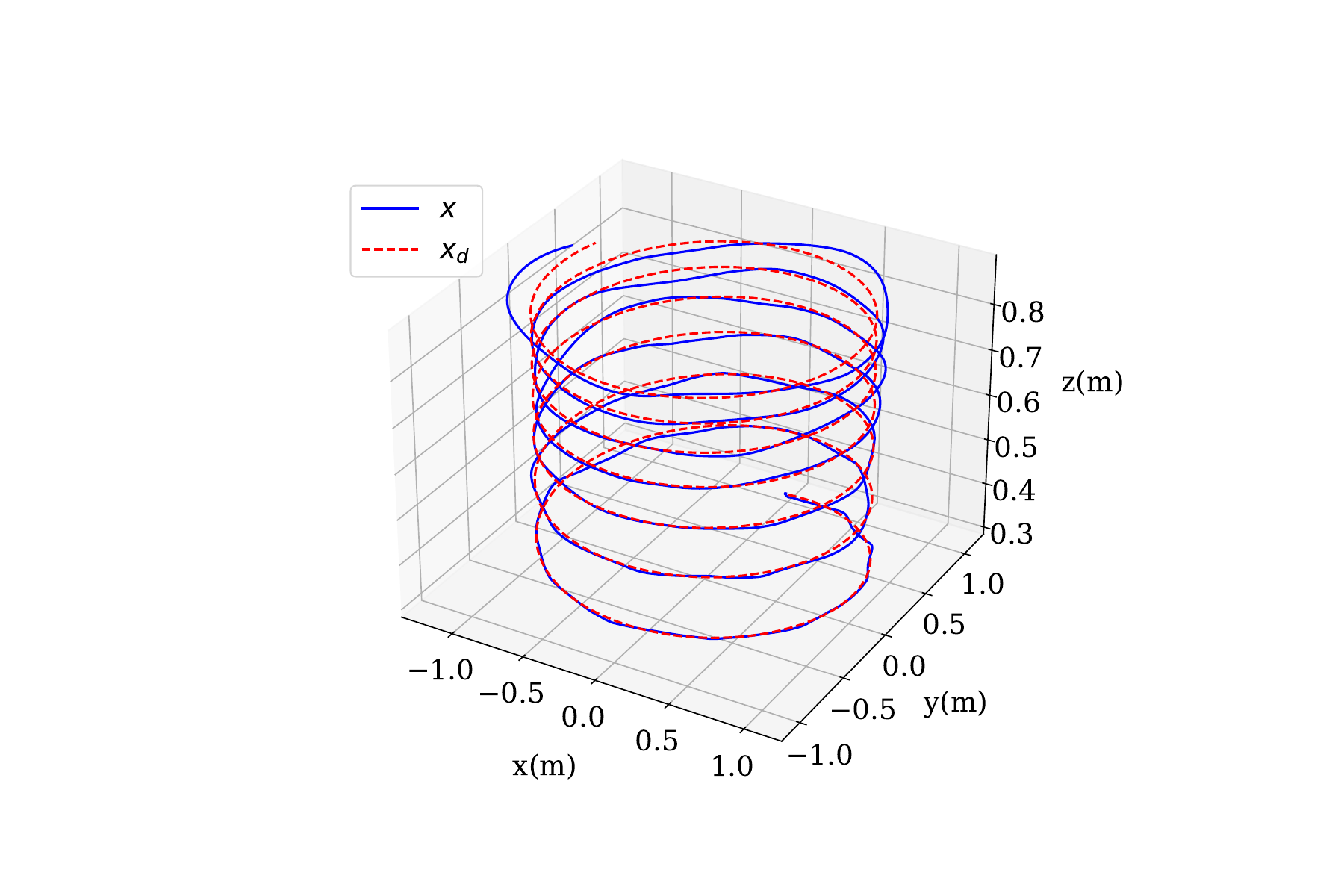}
    \caption{3D desired helix trajectory denoted as $x_d$ and the blimp's real path denoted as $x$.}
    \end{subfigure}
    \begin{subfigure}{1.\linewidth}
    \includegraphics[trim=1.0cm 0.cm 3.cm 0.cm, clip, width=0.95\linewidth]{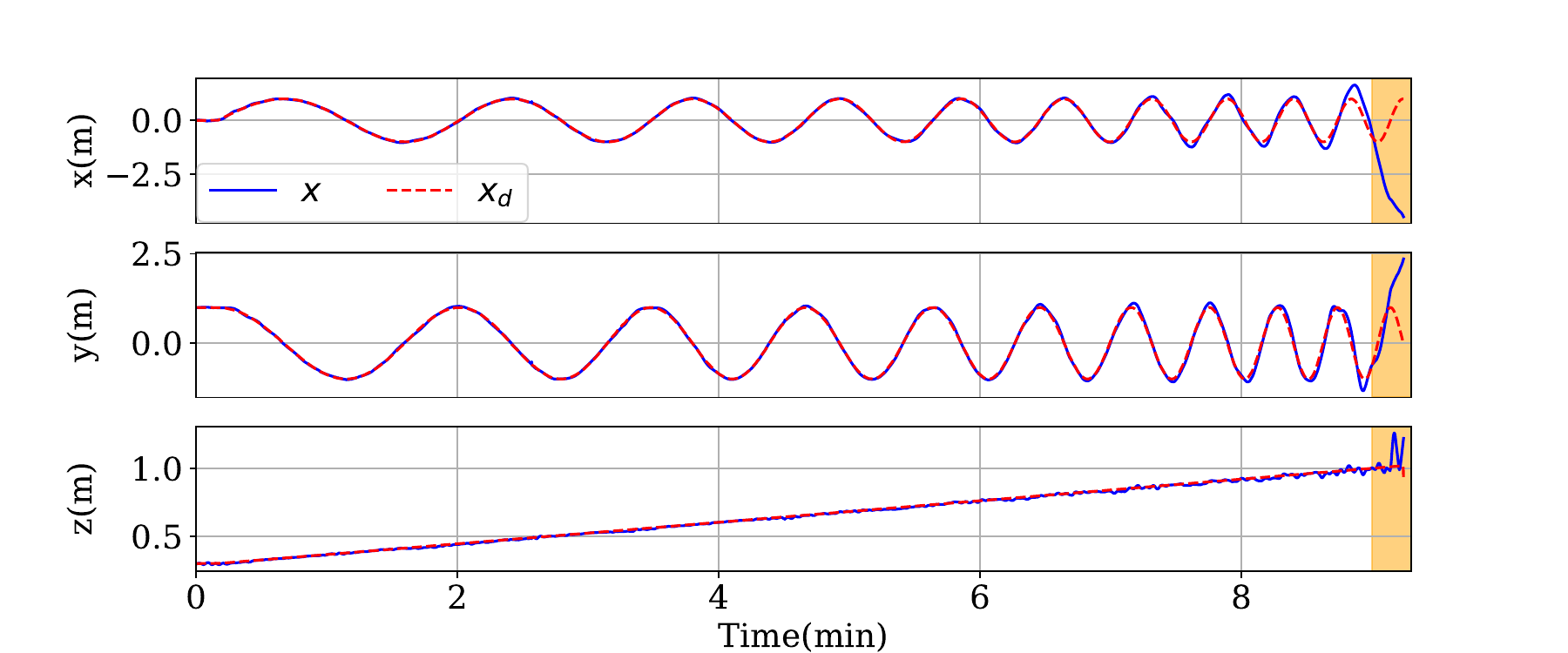}
    \caption{Helix trajectory in $x$-, $y$-, and $z$- axes vs time.}
    \label{fig:helix_b}
    \end{subfigure}
    \caption{The helix trajectory, and the blimp's position readings against the reference. The magnitude of the translational error remains less than $0.25\; m$ and achieves an average of $0.04\; m$.}
    \label{fig:helix}
    \vspace{-1em}
\end{figure}

\subsubsection{Experiment - Tracking a helix trajectory}
% \fixme{Analysis: 600-second experiment duration\\
% Minimum speed is 0.06 m/s and maximum is 0.3 m/s when the blimp loses stability.}
This experiment demonstrates motion in all three axes following the shape of a helix with a radius of $1\; m$, starting from a height of $0.35\; m$ and ending at $1.0\; m$. The ascending speed in $z$-axis remains constant at $0.002\; m/s$ and the speed in the $xy$-plane increases from $0.06\; m/s$ to $0.35\; m/s$. The trajectory is described by linear velocities $\dot x = -\sin{\frac{2\pi}{v(t)}t}$, $\dot y = \cos{\frac{2\pi}{v(t)}t}$, and $\dot z = 0.002$ where $v(t) = 0.06 + 0.000537t$ is the increasing target speed.
The results of this experiment are shown in Fig.~\ref{fig:helix}, with two plots showing the blimp's position relative to the desired. The blimp manages to keep the magnitude of its position error at an average of $0.04\; m$. As the target speed increases, the error remains under $0.25\; m$ before losing stability in the trajectory tracking. In the orange region of Fig.\ref{fig:helix_b}, as the target speed in the $xy$-plane exceeds $0.35\; m/s$, the rotors of the blimp saturates while trying to drive the blimp to follow the desired speed. The combined disturbance caused by the air drag and the input mismaatch between the model and the reality causes the blimp to lose stability.

\subsubsection{Experiment - Colliding with obstacles}

\begin{figure}[t]
    \centering
    % \begin{subfigure}{1.\linewidth}
    % \includegraphics[trim=0.2cm 0cm 0cm 0cm, clip, width=1.\textwidth]{figures/circle_xy.pdf}
    % \caption{Circular Trajectory with low translational error in $xy$-plane.\david{what is this figure trying to show?}}
    % \label{fig:circle_a}
    % \end{subfigure}
    %%
    % \begin{subfigure}{1.\linewidth}
    \includegraphics[trim=1cm 0cm 2cm 0cm, clip, width=.93\linewidth]{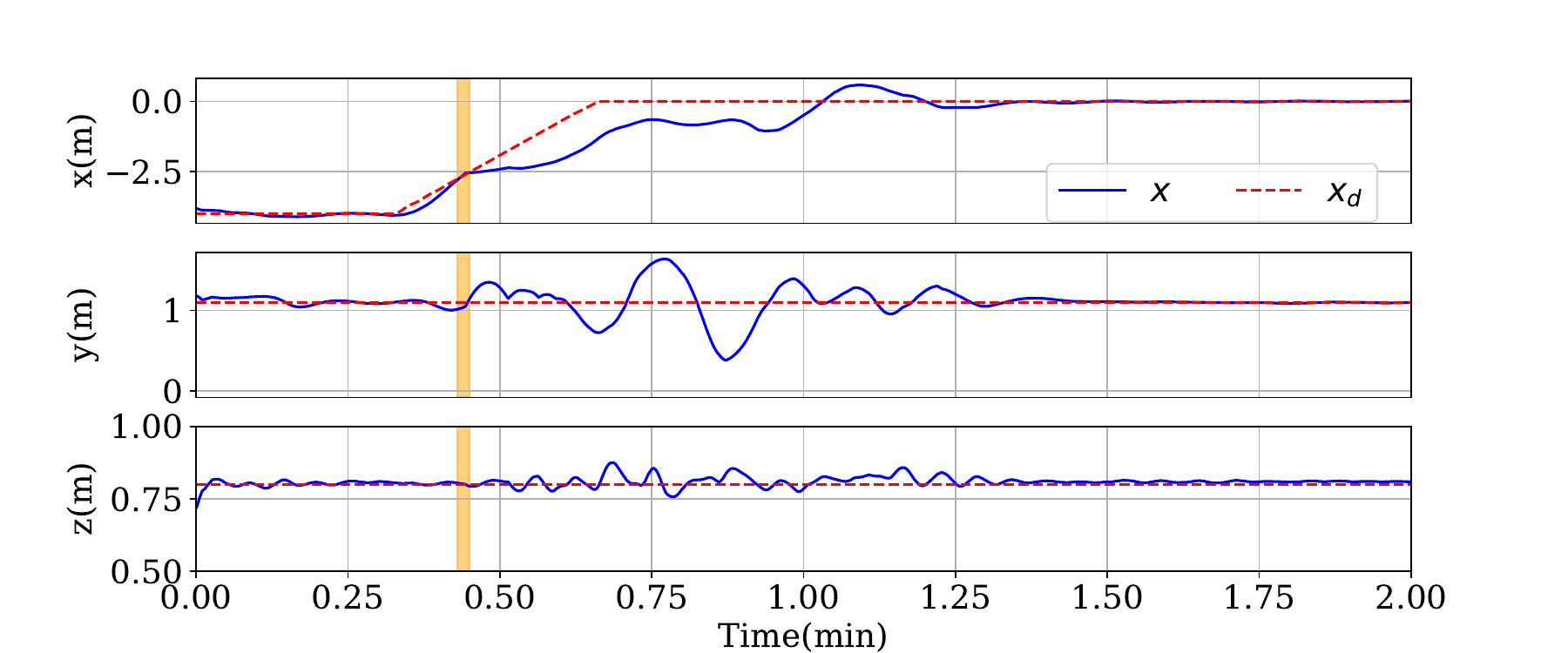}
    % \caption{A box whisker plot showing the magnitude of the position errors against increasing target velocities.}
    % \label{fig:circle_b}
    % \end{subfigure}
    \caption{The blimp's position readings against the reference while following a path and colliding with an obstacle. }
    % The magnitude of the translational error remains less than $0.05$ m and achieves an average of $0.02$ m at a speed of $0.1$ m/s.}
    \label{fig:obstacle} 
    \vspace{-1em}
\end{figure}

We demonstrate the collision-tolerant capabilities of the SBlimp platform. Although the collision recovery is not a focus of this work, the excessive dimensions of the balloon offers the blimp with resilience to collision as the it prevents a direct contact between the actuators and the external object. We direct the blimp towards a window where its path intersected with the frame. The experimental results are presented in Fig.~\ref{fig:obstacle}, which shows that in the orange region of the plot, the SBlimp collided with the window at a speed of $0.25 \ m/s$ and recovered its trajectory within $30\; s$. Despite the collision, the SBlimp could recover and continue its intended trajectory.

%As shown in the video demonstration, 

\section{Conclusion and Future Work}
In this paper, we proposed the design of the SBlimp and presented the planar dynamic model of the blimp. Utilizing the natural pendulum-like stability, we developed a translational motion controller for the miniature robotic blimp and proved its stability in 2-D. The controller allows it to translate regardless of its orientation. We implemented a numerical simulator for the SBlimp in 2-D and constructed a prototype in real-world based on a Crazyflie 2.1 quadrotor. Through simulation, we evaluated factors that affect the performance of the blimp. 
Finally, we demonstrated the effectiveness of our design and the controller in experiments using a real-world prototype, where the blimp follows a series of trajectories of different complexities. 
The blimp's design and control demonstrated a low tracking error and an extended flight time, making it a promising platform for long-term traversal tasks.
% These findings have significant implications for the continued development and deployment of blimp technology. With further advancements in design and control, blimps can become an even more viable option.
%
% In the future, we would like to study the controller stability in 3-D to improve the rigorousness. Moreover, in experiments, we observe that the relatively weak DC motors experience saturation as the speed increases due to the unstructured wind disturbance in the environment. To enable better motion robustness and allow the vehicle to carry additional sensors, we would like to scale up the size of the balloon and upgrade the quadrotor to use brushless motors. 
% In future work, we would like to experiment with different shapes and a deeper study of the air drag on the vehicle's surface.
In future work, we would like to experiment with different shapes of blimps and conduct a deeper study on the effect of the air drag on the blimp's surface to improve its performance at higher speeds.

\bibliographystyle{IEEEtran}
\bibliography{ref}

\end{document}